\documentclass[a4paper,conference]{IEEEtran}
\IEEEoverridecommandlockouts

\usepackage{ifpdf}

\usepackage{color}
\usepackage{xcolor}
\usepackage{cite}

\usepackage{multirow}
\usepackage{booktabs}
\usepackage{diagbox}
\ifCLASSINFOpdf
   \usepackage[pdftex]{graphicx}
\else
   \usepackage[dvips]{graphicx}
\fi
\usepackage{amsmath}
\usepackage{algorithmic}

\usepackage{array}

\ifCLASSOPTIONcompsoc
 \usepackage[caption=false,font=normalsize,labelfont=sf,textfont=sf]{subfig}
\else
 \usepackage[caption=false,font=footnotesize]{subfig}
\fi
\usepackage{fixltx2e}
\usepackage{url}

\begin{document}
%

\title{AutoLC: Search Lightweight and Top-Performing Architecture for Remote Sensing Image Land-Cover Classification}

\author{\IEEEauthorblockN{Chenyu Zheng}
\IEEEauthorblockA{School of Computer Science\\
Wuhan University\\
Email: chenyu.zheng666@gmail.com}
\and
\IEEEauthorblockN{Junjue Wang, Ailong Ma\thanks{Ailong Ma is the corresponding author. This work was supported by National Natural Science Foundation of China under Grant No. 42071350, 42171336, and LIESMARS Special Research Funding.}, Yanfei Zhong}
\IEEEauthorblockA{LIESMARS\\
Wuhan University\\
Email: \{kingdrone, maailong007, zhongyanfei\}@whu.edu.cn}}


%


\maketitle

\begin{abstract}
Land-cover classification has long been a hot and difficult challenge in remote sensing community. With massive High-resolution Remote Sensing (HRS) images available, manually and automatically designed Convolutional Neural Networks (CNNs) have already shown their great latent capacity on HRS land-cover classification in recent years. Especially, the former can achieve better performance while the latter is able to generate lightweight architecture. Unfortunately, they both have shortcomings. On the one hand, because manual CNNs are almost proposed for natural image processing, it becomes very redundant and inefficient to process HRS images. On the other hand, nascent Neural Architecture Search (NAS) techniques for dense prediction tasks are mainly based on encoder-decoder architecture, and just focus on the automatic design of the encoder, which makes it still difficult to recover the refined mapping when confronting complicated HRS scenes.

To overcome their defects and tackle the HRS land-cover classification problems better, we propose AutoLC which combines the advantages of two methods. First, we devise a hierarchical search space and gain the lightweight encoder underlying gradient-based search strategy. Second, we meticulously design a lightweight but top-performing decoder that is adaptive to the searched encoder of itself. Finally, experimental results on the LoveDA land-cover dataset demonstrate that our AutoLC method outperforms the state-of-art manual and automatic methods with much less computational consumption.
\end{abstract}


%
\IEEEpeerreviewmaketitle

\begin{figure*}  
\centering  
\includegraphics[width=0.9 \textwidth]{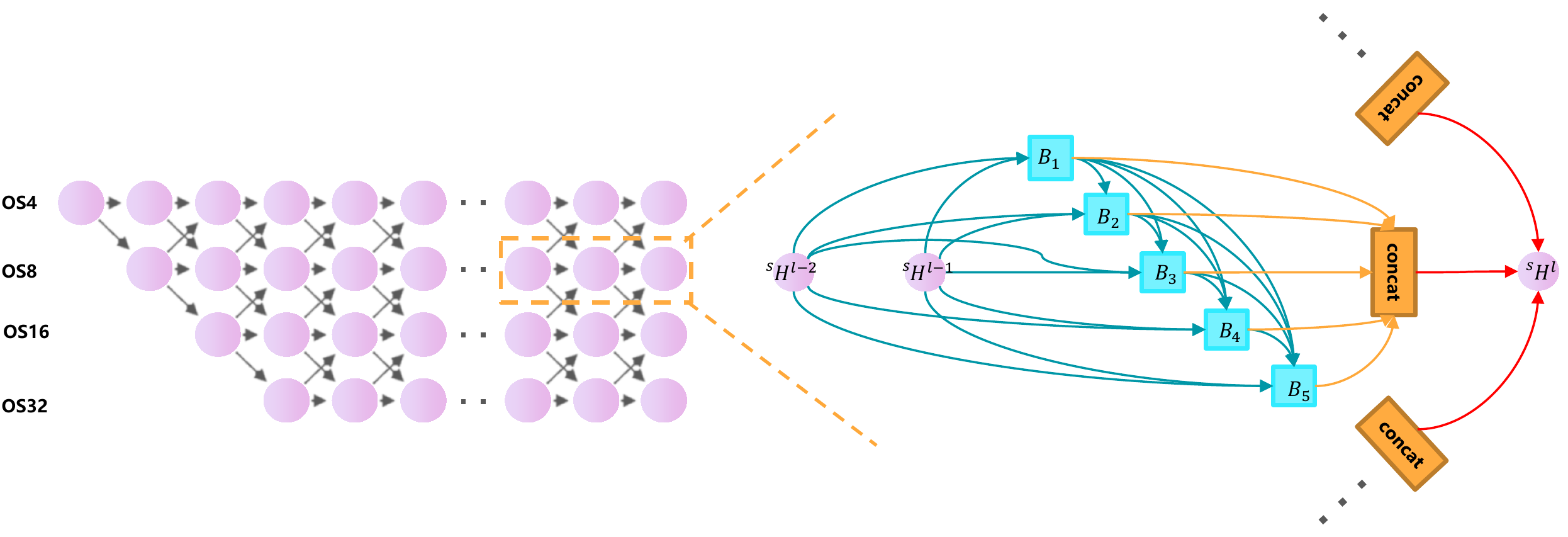}
\caption{Left: Our architecture level search space. Each path along
the pink nodes represents a candidate architecture. The meaning of gray arrows among pink nodes is listed in Table \ref{table:table002}. Right: During the search process, each cell with $5$ blocks is a densely connected structure as described in Sec. \ref{sssec:optim}. Best viewed in color.}  
\label{fig:picture001}  
\end{figure*}

\section{Introduction}
\label{sec:intro}
Land-cover classification, aiming to distinguish High-resolution Remote Sensing (HRS) images with pixel-wise precision, has long been a difficult challenge in remote sensing community. With the advance of earth observation technology, a larger amount of high-resolution remote sensing (HRS) images which contain more detailed spatial information become available. This trend caters to Deep Learning (DL) technique's desire for massive data to learn refined HRS land-cover mapping. However, conventional manual DL techniques\cite{S1, S2, S3, S4, deeplabv3+} are almost proposed for natural images, making it becomes redundant and inefficient to process HRS images. Besides, due to the design of the neural architecture heavily relying on researchers’ prior knowledge and domain experiences, it is difficult for people to jump out of their original thinking paradigm and design a top-performing neural networks with lightweight parameters for HRS land-cover classification.

More recently, in the spirit of AutoML, there has been significant interest in designing neural networks automatically in computer vision community. As an outstanding representative, Neural Architecture Search (NAS) algorithm\cite{rlnas, nasnet} makes it possible to generate neural networks with low consumption according to different tasks, such as image classification\cite{darts} and semantic segmentation\cite{autodeeplab}. In addition, a few NAS algorithms have been proposed for remote sensing recognition tasks at present, including scene classification\cite{scenenet} and land-cover classification\cite{rsnet} .Unfortunately, it is still hard to apply these NAS methods to land-cover classification directly in order to get good performance. We find that nascent NAS techniques for dense image prediction tasks are mainly based on encoder-decoder architectures and just focus on the automatic design of the encoder. Their simple decoders are stiffly connected to the searched encoder and not able to make full use of the rich semantic information from it, so it is still hard for them to recover refined spatial resolution when confronting more complicated HRS scenes\cite{loveda}.

To overcome both approaches' defects and tackle the land-cover classification problems better, we propose AutoLC with automatic encoder and adaptive decoder, which identified architectures that not only achieve the state-of-art performance but also keep lightweight. At last, we evaluate the performance of our method on LoveDA dataset\cite{loveda}, a new public complicated HRS land-cover dataset. The main contributions of our work can be summarized as follows:
\begin{enumerate}
\item We propose a hybrid solution to combine the advantages of manual and automatic methods to tackle HRS land-cover classification problem better.
\item We devise a two-level hierarchical search space and gain a lightweight encoder underlying gradient-based search strategy.
\item We meticulously design a lightweight but top-performing decoder which is adaptive to the searched encoder of itself. This simple decoder efficiently integrates the semantic information from the  searched encoder and do help networks learn finer mapping.
\item Neural networks generated by AutoLC not only achieve state-of-art performance but are also lightweight enough to process massive HRS images efficiently. For instance, it outperforms PSPNet\cite{S3} with $13 \times$ fewer parameters and $7 \times$ less computational consumption.
\end{enumerate}

\section{Related Work}
\label{sec:related work}
\subsection{Manual Deep CNNs for Land-Cover Classification}
In the past few years, manual deep convolutional neural networks (CNNs) based on the fully convolutional neural networks have shown striking improvement over conventional statistical machine learning algorithms\cite{SM,svm} depending on hand-crafted features. Among them, two ways of designing neural architecture for HRS land-cover classification are proved successful. One is the encoder-decoder structure, which is able to gain more sharp object boundaries. FCNs\cite{S1} and UNet\cite{S2} are two remarkable representatives of them, where the former one uses feature maps addition, and the latter one concatenates feature maps. The other way aims to grasp richer contextual information. In order to realize this goal, PSPNet\cite{S3} makes use of pooling operations at different grid scales (PPM), while DeepLabV3\cite{S4} applies several parallel atrous convolution with different rates (ASPP). More recently, DeepLabV3+\cite{deeplabv3+} takes advantage of both approaches and achieves state-of-art performance. Nevertheless, these CNNs are all proposed for dense prediction of natural images, resulting in many redundant parameters and inefficient processing for massive HRS images. Furthermore, even if understanding these excellent design experiences, it still relies on  researchers’ prior knowledge heavily to design perfect networks for HRS recognition problems\cite{FactSeg}.

\subsection{Neural Architecture Search in Computer Vision}
More recently, AutoML technology represented by NAS has received extensive attention from the computer vision community. It aims at designing optimal neural networks according to different tasks, hence excluding researchers' subjective bias and minimizing human hours and efforts. Early papers mainly made use of Reinforcement Learning (RL)\cite{rlnas,nasnet} and evolutionary algorithms (EA)\cite{AmoebaNet} to generate networks. The former train a recurrent neural network (RNN) that represents a policy to generate
a sequence of symbols specifying the CNN architecture, and the latter produce architectures by mutating the best architectures found so far. However, both approaches tend to require considerable computational consumption, usually thousands of GPU days. To reduce crazy computational consumption, ENAS\cite{ENAS} forces child models to share their parameters during the search for architectures, while DARTS\cite{darts} transforms the original discrete search problem to a differentiable optimization problem, and solves it by gradient descent method. Auto-DeepLab\cite{autodeeplab} extends gradient-based search strategy to dense image prediction tasks, but it is based on encoder-decoder architecture and just focuses on the automatic design of the encoder, hence hard to learn refined mapping when tasks become more complex.

\subsection{Neural Architecture Search in Remote Sensing}
Most recently, because mainstream DL models in remote sensing community are derived from natural image processing, where the network architecture is inherited without consideration of the complexity and specificity of HRS images, researchers began to apply NAS techniques to remote sensing recognition tasks. SceneNet\cite{scenenet} proposed a multi-objective EA framework for scene classification. RSNet\cite{rsnet} constructed a multi-task method for both scene classification and land-cover segmentation. Unfortunately, these attempts mainly utilized NAS technology from computer vision field directly. Due to the nascency of NAS, it is still difficult to guarantee that we can generate good enough neural architectures when HSR scenes become complicated.

\section{Proposed AutoLC Framework}
\label{sec:methods}
This section describes our method based on encoder-decoder architecture. For the encoder, we reuse the search space and strategy adopted in \cite{nasnet, autodeeplab} to keep consistent with previous works. For the decoder, we propose a novel lightweight adaptive decoder based on observation and summarization of many excellent design experiences\cite{deeplabv3+, FPN}. It is composed of the following:

\subsection{Automatic Searched Lightweight Encoder}
\label{ssec:encoder}
Inspired by \cite{nasnet,darts,autodeeplab}, we construct a differentiable hierarchical search space including cell level and architecture level. After that gradient-based method is used to find an optimal neural network from the search space.

\subsubsection{Cell Level Search Space}
\label{sssec:cell space}
As it is illustrated in Fig. \ref{fig:picture001}, a \textit{cell} is a small fully convolutional neural module, which is repeated several times to generate the entire neural network. In addition, it is a single directed acyclic graph (DAG) made up of $B$ \textit{blocks}.

Each block consists of two branches, which receive two feature maps as inputs at first. After that, two operations are chosen from the set including all possible filters to transform inputs respectively. Finally, outputs from two paths are added as the final result of the block. Therefore, block $i$ of Cell $l$ can be represented by a 4-tuple $(I_1, I_2, O_1, O_2)$, where $I_1, I_2 \in I_i^l$ mean selected input tensors, and $O_1, O_2 \in O$ are chosen operators in each branch.

The set of all possible input feature maps, $I_i^l$ , is composed of the outputs of the previous two cells $H^{l-1}$ and $H^{l-2}$, and previous blocks’ output in the current cell ${H_1^l, \dots, H_{i-1}^l}$. And the elements of the set $O$ is concluded in Table \ref{table:table001}.

\begin{table}[htb]
\renewcommand{\arraystretch}{1.5}
\centering
\caption{Candidate Operators for Branches of Block}
\begin{tabular}{ccc} 
\toprule
Type                                         & Kernel Size                   & Operation                 \\ 
\hline
\multirow{4}{*}{Convolution}                 & \multirow{2}{*}{$3 \times 3$} & depthwise-separable conv  \\
                                             &                               & atrous conv with rate 2   \\
                                             & \multirow{2}{*}{$5 \times 5$} & depthwise-separable conv  \\
                                             &                               & atrous conv with rate 2   \\ 
\hline
\multicolumn{1}{c}{\multirow{2}{*}{Pooling}} & \multirow{2}{*}{$3 \times 3$} & average pooling           \\
\multicolumn{1}{c}{}                         &                               & max pooling               \\ 
\hline
\multicolumn{1}{c}{\multirow{2}{*}{Others}}  & -                             & skip connection           \\
\multicolumn{1}{c}{}                         & -                             & null connection                     \\
\bottomrule
\end{tabular}
\label{table:table001}  
\end{table}

\subsubsection{Architecture Level Search Space}
\label{sssec:nn space}
The beginning of the architecture is a two-layer CNNs that each reduces the spatial resolution by a factor of 2. After that, as it is displayed in Fig. \ref{fig:picture001} there are a total of $L$-layer cells with different spatial resolutions. The transition forms between two connected cells are shown in Table \ref{table:table002}. And the goal of the search method is then to find an optimal path in this $L$-layer trellis.

\begin{table}[htb]
\renewcommand{\arraystretch}{1.5}
\centering
\caption{Transitions Between Two Cells}
\begin{tabular}{cc} 
\toprule
Type          & Transition                                    \\ 
\hline
$\nearrow$    & Upsample with factor 2                        \\
$\rightarrow$ & Identity                                      \\
$\searrow$    & \multicolumn{1}{l}{Downsample with factor 2}  \\
\bottomrule
\end{tabular}
\label{table:table002}  
\end{table}

\begin{figure*}  
\centering  
\includegraphics[width=0.9 \textwidth]{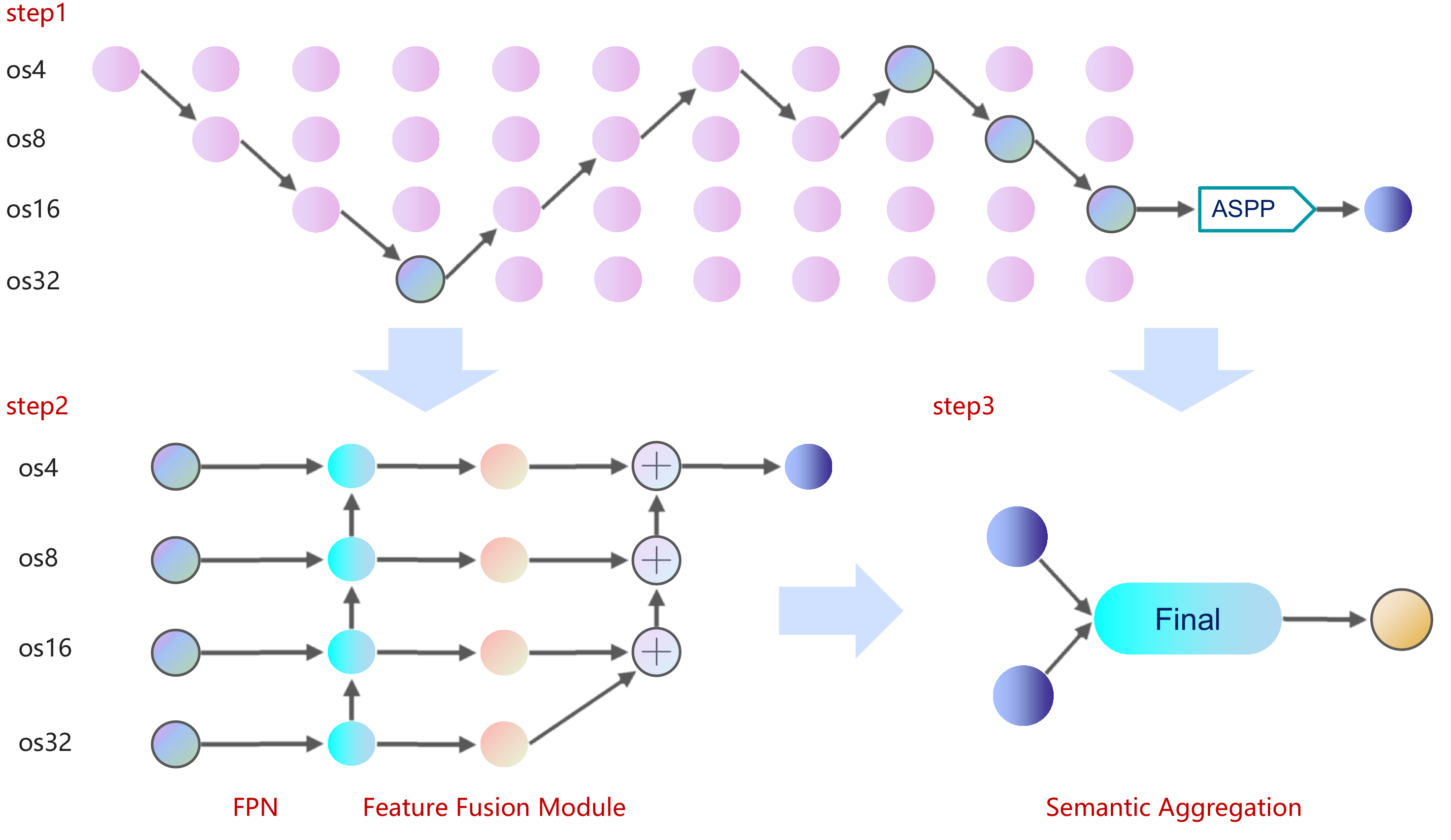}
\caption{Step1: Our searched encoder, and the blue nodes are selected adaptively as inputs of FPN. Step2: Proposed FPN backbone with feature fusion module. Step3: Aggregating semantic information from encoder and adaptive decoder to achieve perfect performance.}  
\label{fig:bestarch}  
\end{figure*}

\subsubsection{Gradient-based Search Strategy}
\label{sssec:optim}
To search for a good architecture efficiently by a gradient-based optimization algorithm, we first reuse continuous relaxation\cite{darts,autodeeplab} to turn the original discrete problem into a continuous optimization problem.

For block $i$ in cell $l$, its output $H^l_i$ is related to all input tensors in $I^l_i$ and all operation elements in $O$, which can be written as follows:
\begin{equation}
H_i^l = \sum_{in \in I_i^l} \sum_{O_k \in O} \alpha_{in \rightarrow i}^k O_k(in)
\end{equation}
where $\alpha_{in \rightarrow i}^k$ are normalized scalars associated with
each operator $O_k \in O$:

\begin{equation}
\sum_{k=1}^{|O|} \alpha_{in \rightarrow i}^k = 1,\quad \forall in, i
\end{equation}

\begin{equation}
\alpha_{in \rightarrow i}^k \ge 0,\quad \forall in, i, k
\end{equation}

As mentioned before, $H^{l-1}$ and $H^{l-2}$ are always included in $I^l_i$, and the result of the cell is just the sum of each block's output, so the computation in cell $l$ can be specified by:

\begin{equation}
H^l = cell(H^{l-1}, H^{l-2}, \alpha)
\end{equation}

In terms of the relaxation in architecture level, we consider the connections among cells. For each layer $l$, it has at most 4 hidden states ${^4H^l, ^8H^l, ^{16}H^l, ^{32}H^l}$, with the upper left superscript indicating the spatial resolution. We define the architecture level update as:

\begin{equation}
\begin{aligned}
^sH^l = \beta_{\frac{s}{2} \rightarrow s}^l cell(^{\frac{s}{2}}H^{l-1}, ^sH^{l-2}, \alpha) \\  +  \beta_{s \rightarrow s}^l cell(^sH^{l-1}, ^sH^{l-2}, \alpha) \\
 +  \beta_{2s \rightarrow s}^l cell(^{2s}H^{l-1}, ^sH^{l-2}, \alpha)
\end{aligned}
\end{equation}
where $s \in \{4, 8, 16, 32\}$ and $l \in \{1, \dots ,L\}$. Different from Auto-DeepLab\cite{autodeeplab},the scalars are normalized as:

\begin{equation}
\beta_{\frac{s}{2} \rightarrow s}^l +  \beta_{s \rightarrow s}^l + \beta_{2s \rightarrow s}^l = 1
\end{equation}

\begin{equation}
\beta_{\frac{s}{2} \rightarrow s}^l,\quad \beta_{s \rightarrow s}^l,\quad \beta_{2s \rightarrow s}^l \ge 0
\end{equation}

After applying continuous relaxing to the original problem, we can then regard it as a two-level continuous optimization problem. We separate the train set into two disjoint sets $trainA$ and $trainB$ at first. Then we make use of the first-order optimization method in \cite{darts} to find an optimal neural network. The updating process within a mini-batch is as follows.
\begin{enumerate}
\item $w \leftarrow w - \eta_m \nabla_w \mathcal L_{trainA}(w, \alpha, \beta)$
\item $(\alpha, \beta) \leftarrow (\alpha, \beta) - \eta_a \nabla_{\alpha,\beta} \mathcal L_{trainB}(w, \alpha, \beta)$
\end{enumerate}
where the loss function $\mathcal L$ is the cross-entropy calculated on the mini-batch. And the $\eta_m$ and $\eta_a$ are learning rates for updating model weights and architecture parameters $(\alpha, \beta)$ respectively.

When optimization ends, an optimal architecture can be decoded from the entire search space. For cell level, we first choose the 2 strongest predecessors for each block and then select the most likely operator. At last, we recover architecture-level connections by using a dynamical programming algorithm\cite{dp} to find the path with the maximum probability from start to end.

\begin{figure}[htb] 
\centering  
\includegraphics[width=0.45\textwidth]{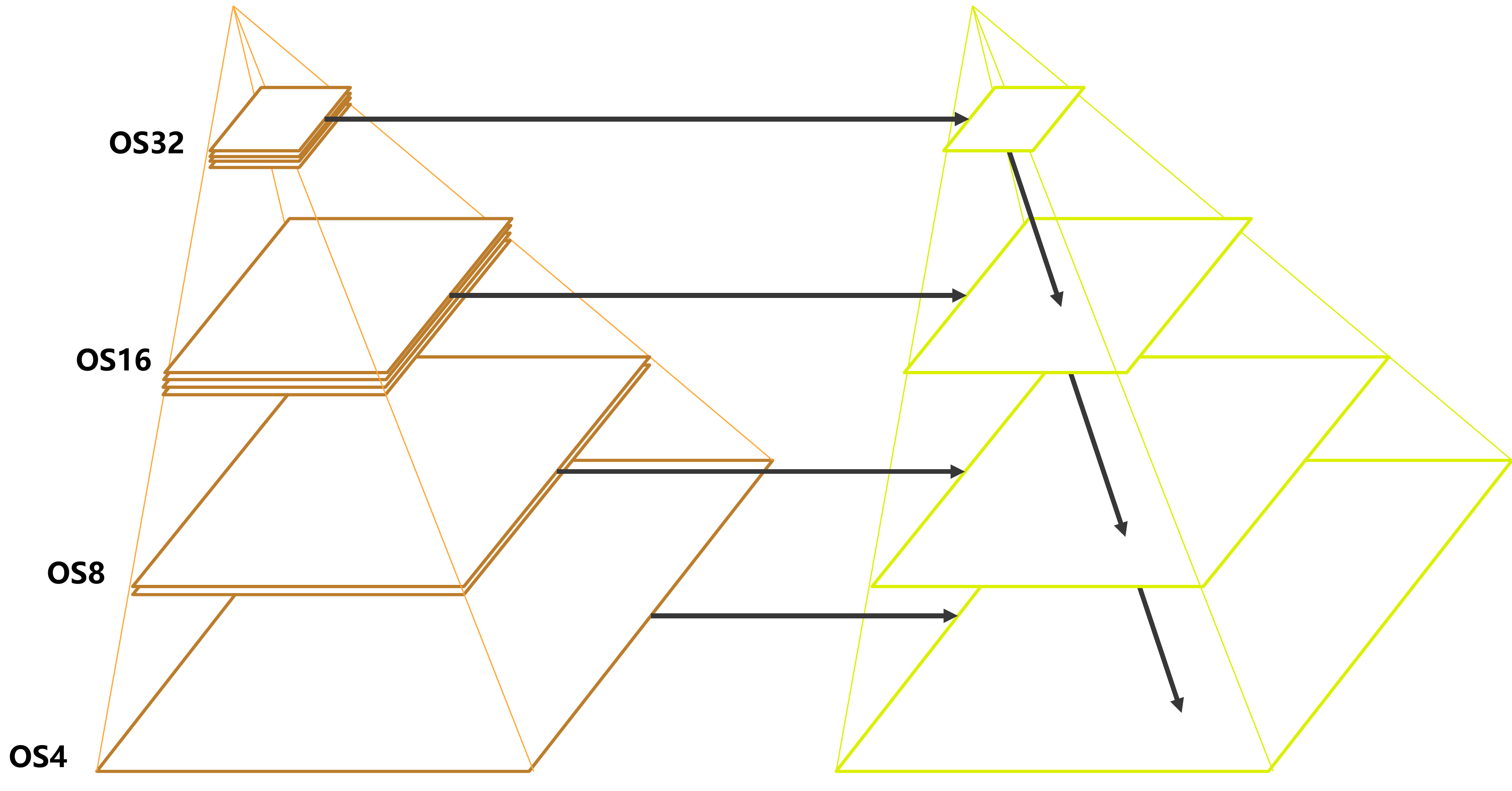}  
\caption{Feature Pyramid Network backbone, which has been widely used in object detection for extracting rich multi-scale features.}
\label{fig:fpn}  
\end{figure}

\subsection{Adaptive Lightweight Decoder}
\label{ssec:decoder}
Infant NAS methods just focus on the automatic search for the encoder, resulting in its poor performance on complicated tasks. Inspired by previous excellent experiences, we design an adaptive decoder inspired by Feature Pyramid Network (FPN)\cite{FPN} to solve land-cover classification problems better. It is lightweight and top-performing.

\subsubsection{Adaptive Feature Pyramid Module}
\label{sssec:fpn}

After applying the NAS algorithm mentioned before, we get an encoder whose path may pass through multiple spatial resolutions. For each spatial resolution $s \in \{4,8,16,32\}$, if there exists $^sH^l$ in the finally searched encoder, $argmax_l\ ^sH^l$ is chosen automatically as the input feature maps for the next FPN backbone, which can be implemented as an adaptive procedure. Afterward FPN processes these selected features with different resolutions, it adds a light top-down pathway with lateral connections, see Fig. \ref{fig:fpn}. The top-down pathway starts from the one with minimal resolution in input tensors and progressively upsamples it while adding in transformed versions of higher-resolution features from the bottom-up pathway. As result, FPN generates a pyramid, where each pyramid level has the same channel dimension, which can be denoted as a hyper-parameter $dim$.

\subsubsection{Feature Fusion Module}
\label{sssec:SSB}
\begin{figure}[htb] 
\centering  
\includegraphics[width=0.45\textwidth]{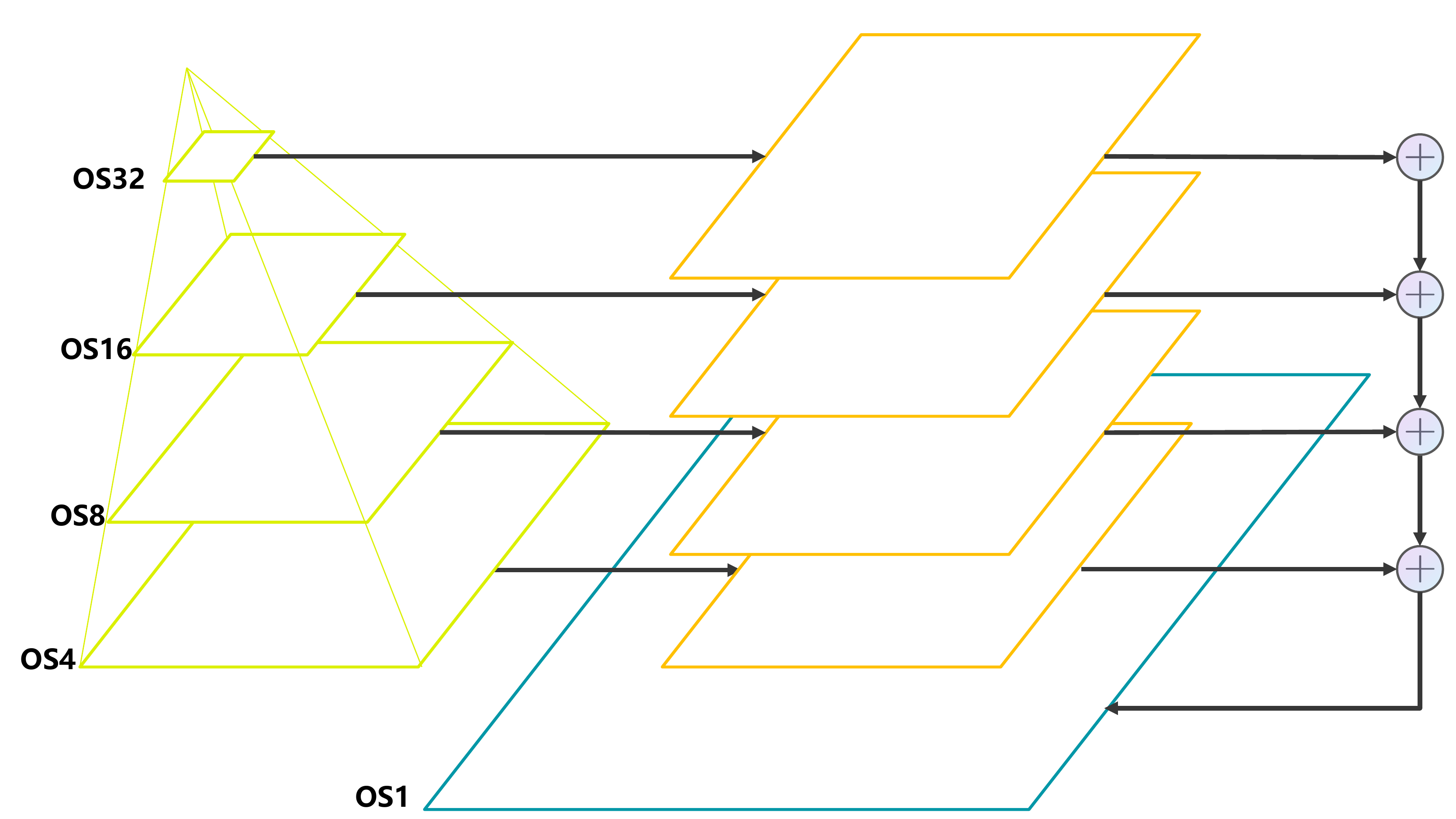}  
\caption{Feature Fusion Module. Each FPN feature map is upsampled until it reaches $os4$, these outputs are then summed and finally transformed into a output with the size of original input.}
\label{fig:adecoder}  
\end{figure}

To generate the semantic segmentation output from the FPN features, we use feature fusion module based on the semantic segmentation branch in \cite{FPN} to merge the information from all levels of the FPN pyramid into a single output. It is illustrated in detail in Fig. \ref{fig:adecoder}. Starting from the deepest FPN level to the shallowest level, we perform several upsampling stages to yield a feature map at scale $1/4$ in each level, where each upsampling stage consists of $3\times3$ convolution, batch norm\cite{bn}, ReLU, and $2 \times$ bilinear upsampling. The result is the sum of the set of feature maps at the same $1/4$ scale.

\subsubsection{Semantic Aggregation Module}
\label{sssec:AAM}
To make the best of the power of encoder-decoder structure and rich contextual information, we combine the outputs of the feature fusion module and ASPP module of the encoder. After that, a final  $3\times3$ convolution,  $1\times1$ convolution, 4 bilinear upsampling, and softmax are used to generate the prediction for per pixel at the original image resolution. It is displayed in Fig. \ref{fig:bestarch}
.

\section{Experiments}
\label{sec:experiments}
In this section, we report our architecture search implementation details as well as the searched encoder. After that, we report and compare our semantic segmentation results on the LoveDA dataset with state-of-art manual and automatic architectures.

\subsection{Dataset Description}
\label{ssec:dataset}
The LoveDA dataset\cite{loveda} contains 5987 HSR ($1024\times1024$) images with 166768 annotated objects from three different cities. Compared to the existing datasets, the LoveDA dataset encompasses two domains (urban and rural), which also brings considerable challenges due to multi-scale objects, complex background samples, and inconsistent class distributions. It requires a much more difficult land-cover mapping than previous datasets.

\begin{table*}[htb]
\renewcommand{\arraystretch}{1.6}
\caption{Performance of the Reference and Our Models on the LoveDA Validation Set}
\centering
\begin{tabular}{cccccccc} 
\toprule
Type                       & Model                 & Params (M)  & FLOPs (G)    & Memory (GB)    & MAdd (T)    & MemR+W (GB)  & mIOU(\%)  \\ 
\hline
\multirow{5}{*}{Mannual}   & FCN8s\cite{S1}                 & 15.31  & 322.1  & 2.28 & 0.64  & 4.55  & 43.14     \\
                           & UNet\cite{S2}                  & 17.27  & 642.9  & 6.10    & 1.28   & 11.85 & 43.27     \\
                           & PSPNet\cite{S3}                & 46.59  & 709.4  & 5.31 & 1.42   & 10.51 & 43.73     \\
                           & DeepLabV3\cite{S4}            & 39.64  & 655.7   & 5.49 & 1.31   & 11.13 & 45.11     \\
                           & DeepLabV3+\cite{deeplabv3+}            & 26.68 & 147.2  & 3.22 & 0.29 & 6.36  & 45.14     \\ 
\hline
\multirow{2}{*}{Automatic} & Auto-DeepLab\cite{autodeeplab}                & 5.30      & 101.2       & 1.94        & 0.20      & 3.83      & 41.29        \\
                           & RSNet\cite{rsnet} & \textbf{3.22}      & 118.3       & 8.93        & 0.24     & 15.4      &  40.71   \\ 
\hline
\multirow{4}{*}{Hybrid (Ours)}    & L10\_F10\_dim64      & 3.68   & \textbf{91.1} & \textbf{1.40} & \textbf{0.18} & \textbf{2.54}  & 44.57     \\ 
                           & L10\_F10\_dim128      & 4.28   & 102.2 & 1.51 & 0.20 & 2.62  & 44.64     \\
                           & L10\_F20\_dim128      & 6.81   & 123.3 & 2.26 & 0.25 & 4.18  & 45.24     \\
                           & L10\_F30\_dim128      & 9.81   & 151.2 & 3.01 & 0.30 & 5.74  & \textbf{45.26}     \\
\bottomrule
\end{tabular}
\label{table:miou}  
\end{table*}

\subsection{Architecture Search  Details}
\label{ssec:searchdetials}
Due to the constrain of GPU memory, we consider a total of $L = 10$ layers in the network, and $B = 5$ blocks in a cell. Every node with downsample rate $s$ has $\frac{BFs}{4}$ output channels, where $F$ is the hyper-parameter controlling the model size. The same as \cite{autodeeplab}, we set $F = 8$ during the architecture search.

We conduct neural architecture search on the LoveDA dataset. More specifically, we use $321\times321$ random images cropped from half-scaled ($512\times512$) images in the train set. In addition, we randomly select half of the images in train fine as $trainA$, and the others as $trainB$. The disjoint set partition is to avoid the architecture searched overfitting the train set.

\begin{figure}[htb] 
\centering  
\includegraphics[width=0.48\textwidth]{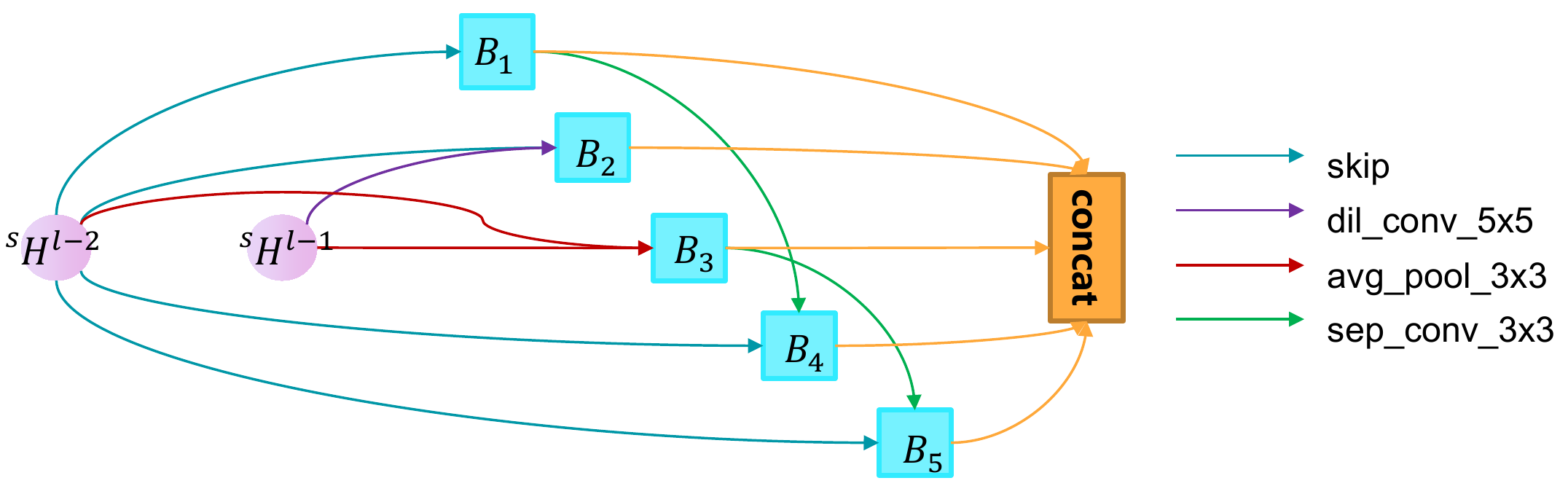}  
\caption{Cell level architecture searched on the LoveDA dataset.}
\label{fig:cell}  
\end{figure}

The architecture search algorithm is executed for a total of $60$ epochs. Keeping consistent with \cite{autodeeplab}, when updating network weights $w$, we use SGD optimizer with momentum $0.9$, cosine learning rate that decays from $0.025$ to $0.001$, and weight decay $0.0003$. The initial values of $\alpha, \beta$ before normalized are sampled from a standard Gaussian times $0.001$. Afterward they are optimized using Adam optimizer\cite{adam} with learning rate $0.003$ and weight decay $0.001$. We begin optimizing architecture parameters after $30$ epochs. The entire architecture search optimization takes about $8$ days on one TITAN X GPU. Fig. \ref{fig:cell} shows that the cell level architecture of the best found network.

Fig. \ref{fig:bestarch} visualizes the best architecture we found. In terms of architecture level, higher resolution is preferred at the end to narrow the gap of resolution between the feature before upsampling and the original HRS image. Besides, our found architecture is similar to Stacked Hourglass\cite{SH}, which suggests that transitions among nodes with different resolutions may be important for remote sensing tasks. In terms of cell level structure, what really surprised us is that more than half of the selected operations are pooling, skip connection, even null connection, while convolution operators only account for a third. Fortunately, subsequent experiments proved this architecture is not only lightweight but also top-performing for HRS image land-cover classification. In reality, it is almost impossible for people to design such a seemingly poor model to solve recognition tasks, so the significance of the NAS method is proved again.

\subsection{Land-cover Segmentation Result}
\label{ssec:lcans}
We evaluate the performance of our searched best encoder and adaptive decoder on the LoveDA dataset. Results proved that AutoLC is able to find lightweight and perfect neural networks for complex land-cover classification problems.

During training, we use a polynomial learning rate schedule with an initial learning rate 0.05, and large crop size ($521 \times 521$). The models are trained from scratch with $95k$ iterations while the batch size is 8. Furthermore, we adopt the warm-up technique during the first 5k iterations.

In Table \ref{table:miou}, fixing $dim$ as 128, we compare the accuracy and efficiency of our AutoLC with other manual and automatic semantic segmentation benchmarks on the LoveDA validation set. Specially, we adopt mIOU as our evaluation metric for accuracy. Besides, our evaluation metrics for model efficiency include a total number of network parameters (Params), amount of floating-point arithmetics (FLOPs), amount of multiply-adds (MAdd), memory usage, and amount of reading and writing memory (MemR+W). The input for efficiency analysis is a $1024 \times 1024$ HSR image.

Experiment results show that AutoLC outperforms not only manual neural networks but also automatically searched architectures. For one thing, compared to classical manual networks, AutoLC reaches state-of-art performance with a much more efficient setting. It is worthwhile to note that when $F = 20$ and $dim = 128$, AutoLC outperforms all benchmarks above with $11\times$ fewer parameters and $7\times$ fewer FLOPs than PSPNet\cite{S3}. For another, AutoLC gains great improvement on accuracy and less computational consumption than pure NAS algorithms\cite{autodeeplab, rsnet} with slightly more parameters.

\begin{table}
\renewcommand{\arraystretch}{1.6}
\caption{Results of Sensitivity Analysis for Hyper-parameters}
\centering
\begin{tabular}{c|ccc} 
\toprule
\diagbox{F}{mIOU(\%)}{dim} & 64     & 128    & 256     \\ 
\hline
10                     & 0.4457 & 0.4464 & 0.4321  \\
20                     & 0.4453 & 0.4524 & 0.4523  \\
30                     & 0.4358 & 0.4526 & 0.4466  \\
\bottomrule
\end{tabular}
\label{paramsanalysis}
\end{table}

\subsection{ Parameter Sensitivity Analysis}
\label{ssec:parasa}

In order to investigate the effect of different hyper-parameters on our method's performance, we conduct sensitivity analysis for parameters controlling model capacity ($F$) and the number of channels of FPN's output ($dim$). And the results are displayed in Table \ref{paramsanalysis}.

As we can observe, results on different conditions are competitive enough, so the performance of our method, in general, is not very sensitive to hyper-parameters $F$ and $dim$. More concretely, we can summarize some practical experiences for AutoLC as follows.

\begin{enumerate}
\item AutoLC is very suitable for efficient prediction in extreme situation. As it is illustrated in Table \ref{table:miou} and \ref{paramsanalysis}, when $(F, dim) = (10,64)$, it defeats most of benchmarks listed in the Table \ref{table:miou} with extraordinarily few parameters ($1/13$ of PSPNet) and computational consumption ($1/7$ FLOPs of PSPNet).
\item In order to prevent the performance of the model become bad, the ratio of the $F$ and $dim$ is supposed to be moderate. Small $F$ with big $dim$ or the reverse side is not good. As we can see, when $(F, dim) = (10,256)$ and $(F, dim) = (30,64)$, it reach the two worst performance even with more model capacity than $(F, dim) = (10,64)$. We think an unbalanced relationship between two hyper-parameters makes it more difficult for neural networks to converge.
\item Based on the harmonious proportion of hyper-parameters, it does not mean that larger capacity leads to better performance. For instance, performance condition on $(F, dim) = (30,256)$ worse than $(F, dim) = (20,128)$. It should be determined by the complexity of the problem.
\end{enumerate}

\section{Conclusion}
\label{sec:conclusion}
In this paper, we present one of the few attempts to combine the advantages of both manual and automatic architectures to gain more refined HRS land-cover mapping. Our proposed AutoLC is based on the encoder-decoder architecture. More specially, we adopt the NAS algorithm to generate an efficient encoder and meticulously design a lightweight but top-performing decoder. The architectures induced by our method are evaluated by training on the LoveDA dataset from scratch. On LoveDA, AutoLC outperforms all listed previous state-of-art benchmarks with $11\times$ fewer parameters and $7\times$ less computational consumption than PSPNet\cite{S3}.

For future work, for one thing, our AutoLC can be extended to more efficient dense prediction tasks. For another, more advanced NAS search spaces and optimization methods can be proposed to generate better architectures.






%



\bibliographystyle{IEEEtran}
\bibliography{SS, NAS}

\end{document}